\documentclass{article}
\usepackage{spconf,amsmath,graphicx, hyperref}
\usepackage{amssymb}
\usepackage{multirow}
\usepackage{booktabs}
\usepackage{rotating}
\usepackage{tablefootnote}

\title{SDGF: Fusing Static and Multi-Scale Dynamic Correlations for Multivariate Time Series Forecasting}
\name{Shaoxun Wang$^{*1}$ \thanks{*Corresponding author: Xingjun Zhang and Shaoxun Wang. This work is partially supported by National Natural Science Foundation of China 62372366.}, Xingjun Zhang$^{*1}$, Qianyang Li$^{1}$, Jiawei Cao$^{1}$, Zhendong Tan$^{1}$ }

\address{
    $^{1}$School of Computer Science and Technology, Xi'an Jiaotong University, Xi'an, China\\
    \{shaoxunwang, liqianyang, 772316639\}@stu.xjtu.edu.cn, xjzhang@xjtu.edu.cn 
}
\begin{document}
%
\maketitle
\begin{abstract}
Accurate multivariate time series forecasting hinges on inter-series correlations, which often evolve in complex ways across different temporal scales. Existing methods are limited in modeling these multi-scale dependencies and struggle to capture their intricate and evolving nature. To address this challenge, this paper proposes a novel Static-Dynamic Graph Fusion network (SDGF), whose core lies in capturing multi-scale inter-series correlations through a dual-path graph structure learning approach. Specifically, the model utilizes a static graph based on prior knowledge to anchor long-term, stable dependencies, while concurrently employing Multi-level Wavelet Decomposition to extract multi-scale features for constructing an adaptively learned dynamic graph to capture associations at different scales. We design an attention-gated module to fuse these two complementary sources of information intelligently, and a multi-kernel dilated convolutional network is then used to deepen the understanding of temporal patterns. Comprehensive experiments on multiple widely used real-world benchmark datasets demonstrate the effectiveness of our proposed model. Code is available at \href{https://github.com/shaoxun6033/SDGFNet}{https://github.com/shaoxun6033/SDGFNet}.
\end{abstract}

\section{Introduction}
\label{sec:intro}

Time Series Forecasting (TSF) is a crucial research topic in the field of machine learning. It demonstrates significant application value across numerous critical domains, including financial market analysis~\cite{lu2024trnn}, smart energy management~\cite{salman2024hybrid}, meteorological disaster early warning~\cite{wang2021real}, and intelligent transportation planning \cite{ma2021short}. Accurate forecasting capability forms the cornerstone for scientific decision-making, optimal resource allocation, and proactive risk management in modern society.

In multivariate time series forecasting, it is equally important to accurately capture the temporal dynamics within individual sequences (intra-series dependencies) and to effectively model the interactions among different sequences (inter-series dependencies) \cite{Cai2024}. In recent years, advanced architectures—from RNNs \cite{amalou2022multivariate}, LSTM \cite{wang2023dafa}, TimeMixer \cite{wang2024timemixer}, and DLinear \cite{Zeng2023} to Transformer-based models such as Informer \cite{zhou2021informer}, Autoformer \cite{wu2021autoformer}, and PatchTST \cite{nie2022time}—have demonstrated remarkable success in capturing long-range temporal dependencies and intra-series patterns. Meanwhile, Graph Neural Networks (GNNs) \cite{kipf2016semi} have shown significant advantages in handling graph-structured data due to their permutation invariance, local connectivity, and compositionality. By viewing multivariate time series from a graph perspective, GNNs provide a natural and effective framework for modeling complex inter-series dependencies. However, existing GNN-based methods exhibit notable limitations: they either heavily rely on a predefined, prior-knowledge-based graph \cite{yu2017spatio}, which is unrealistic in many general scenarios, or—even when adopting adaptive graph learning—typically produce a single global static graph \cite{li2017diffusion}. This “one-size-fits-all” approach overlooks the crucial fact that inter-variable correlations may exhibit markedly different patterns across multiple temporal scales. Although some efforts have attempted to address this issue \cite{Cai2024, wu2020connecting}, they still struggle to fully capture the complex and dynamically evolving inter-series dependencies.

To address these challenges, we introduce multi-level wavelet decomposition (MWD) to decompose the original multivariate time series into sub-sequences at different frequencies and temporal scales, thereby capturing multi-granularity inter-variable relationships. Based on this, we propose the static-dynamic graph fusion (SDGF) Network, which models long-term stable dependencies via a global static graph and learns dedicated dynamic graphs for each wavelet scale to capture scale-dependent relations.

Our main contributions are: (1) We propose the SDGF Network, which simultaneously models static and dynamic scale-dependent inter-series relations. (2) We introduce an attention-gated fusion mechanism to effectively integrate multi-scale dynamic graph convolutions with a global static graph. (3) We validate the effectiveness of our model on multiple benchmark datasets, showing superior predictive performance over existing state-of-the-art methods.

\section{RELATED WORK}
\label{sec:prior}
Deep learning-based time series forecasting has become the mainstream in research and applications, with MLPs and Transformers serving as the two dominant frameworks. MLP-based methods are favored for their simplicity and interpretability, such as TimeMixer \cite{wang2024timemixer}, DLinear \cite{Zeng2023}, and TimesNet \cite{wu2022timesnet}. Transformer-based methods, exemplified by Informer \cite{zhou2021informer}, Autoformer \cite{wu2021autoformer}, and TimeXer \cite{wang2024timexer}, exploit self-attention to capture sequence dependencies, often combined with decomposition or Fourier transforms to enhance temporal modeling.

GNNs are specifically designed for graph-structured data, making them well-suited for modeling complex interactions and learning local patterns in time series. They were initially applied to spatio-temporal traffic forecasting \cite{li2017diffusion} and skeleton-based action recognition \cite{shi2019skeleton}. Later, MTGNN \cite{wu2020connecting} explored a learnable graph structure, while MSGNet \cite{Cai2024} further advanced this direction by integrating frequency-domain analysis with adaptive graph convolutions, effectively capturing complex dependencies in multivariate time series data. However, these approaches typically rely on a limited number of graph structures or scales, and fail to incorporate global static structural relations.

\section{METHOD}
\label{sec:majhead}

\subsection{Problem Statement}
\label{ssec:subhead}
Given a historical multivariate time series $X_{in} \in \mathbb{R}^{L \times N}$, where $L$ denotes the length of the historical window and $N$ is the number of variables, the task is to predict the future sequence $\hat{X}_{out} \in \mathbb{R}^{T \times N}$ for the next $T$ steps. To capture inter-variable dependencies, we employ wavelet decomposition to project the input series into multiple scales, and learn graph structures at each scale, yielding a set of adjacency matrices $\{A_1, A_2, \dots, A_k\}$ with $A_j \in \mathbb{R}^{N \times N}$. Finally, the forecasting function integrates the input sequence with the multi-scale graph structures to produce predictions as $\hat{X}_{out} = f(X_{in}, \{A_j\}_{j=1}^k)$.

\subsection{Framework}
\label{ssec:Framework}

\begin{figure}[htb]  
  \centering
  \includegraphics[width=8.5cm]{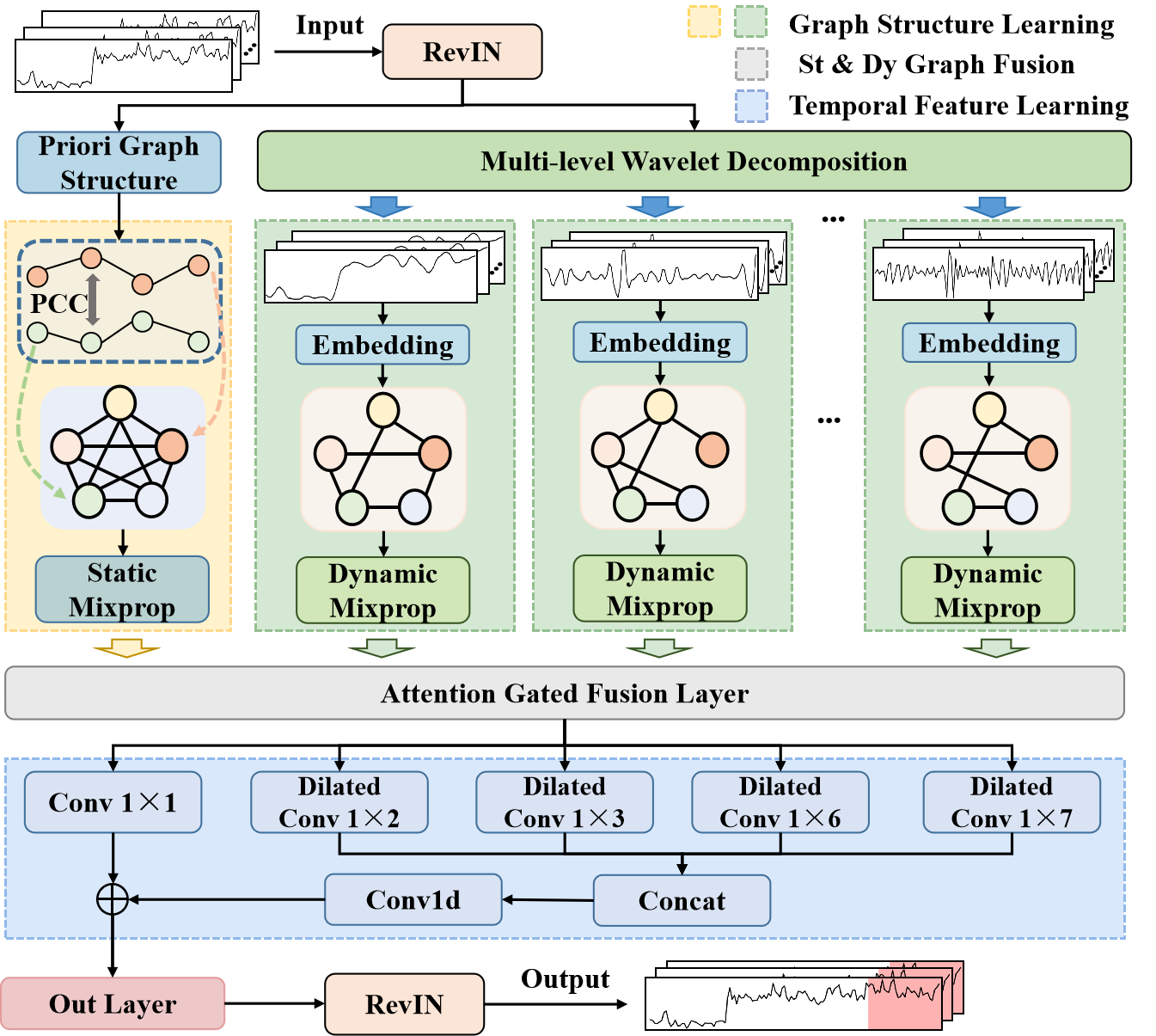}  
  \caption{Overall architecture of SDGF Network.}
  \label{fig:framework}
\end{figure}

Our proposed \textbf{SDGF} Network aims to capture complex inter-series correlations at different temporal scales through graph structure learning and enhance predictive accuracy by fusing dynamic and static graph structures. The overall architecture is illustrated in Fig.~\ref{fig:framework}, and it consists of three core modules: (1) Graph Structure Learning: The input sequences are first normalized using RevIN normalization \cite{kim2021reversible} to eliminate scale differences across series. Subsequently, multi-level wavelet decomposition (MWD) is applied to extract multi-scale features, providing diverse frequency components for graph structure learning. The static graph structure (Priori Graph Structure) leverages prior knowledge or statistical correlations(Pearson Correlation Coefficient, PCC) to construct a fixed inter-series relationship graph, capturing long-term stable dependencies. In contrast, the dynamic graph structure adaptively learns variable inter-series relationships based on multi-scale features, capturing short-term or evolving dependencies. (2) Attention Gated Fusion: To effectively integrate static and dynamic graph convolution outputs, we employ an attention gated fusion module, which adaptively combines these features, yielding more accurate and flexible inter-series representations. (3) Temporal Feature Learning: The fused inter-series features are processed through multi-kernel dilated convolutions to extract high-level temporal representations, capturing both short- and long-term dependencies across multiple time scales. Finally, the learned features are fed into a lightweight MLP-based Output layer for nonlinear mapping to produce the final forecasts. In summary, SDGF Network leverages multi-scale graph structure learning, attention-based fusion, and advanced temporal feature extraction to provide an efficient and scalable framework for multivariate time series forecasting.

\subsection{Graph Structure Learning}
\label{ssec:Graph}
To eliminate distribution shift and scale differences, we first apply Reversible Instance Normalization (RevIN) to the input sequences, 
$\mathbf{X}_{\text{norm}} = \text{RevIN}(\mathbf{X}, \text{mode}=\text{'norm'})$, 
where RevIN normalizes each instance by subtracting its mean and dividing by its standard deviation. Learnable affine parameters $(\gamma, \beta)$ ensure reversibility, allowing the predicted outputs to be mapped back to the original scale during inference.

Static Graph Structure (Prior Graph): The static graph captures long-term stable inter-series dependencies. Given the normalized input $\mathbf{X}_{\text{norm}} \in \mathbb{R}^{B \times L \times N}$, where $B$ is the batch size, $L$ is the sequence length, and $N$ is the number of variables, the adjacency matrix $\mathbf{A} \in \mathbb{R}^{N \times N}$ is computed based on the Pearson Correlation Coefficient (PCC):
\begin{equation}
\mathbf{A}_{nm} = \frac{\text{Cov}(\mathbf{X}_{\text{norm},:,n}, \mathbf{X}_{\text{norm},:,m})}{\sigma_n \sigma_m + \epsilon}
\end{equation}
where $\text{Cov}(\cdot,\cdot)$ is the covariance, $\sigma_n$ and $\sigma_m$ are the standard deviations of sequences $n$ and $m$, and $\epsilon=10^{-9}$ prevents division by zero. Averaging over the batch dimension and applying ReLU followed by softmax ensures non-negativity and normalization:
\begin{equation}
\mathbf{A} = \text{Softmax}\big(\text{ReLU}(\text{Mean}_B(\mathbf{A}))\big)
\end{equation}

The static graph convolution propagation is formulated as:
\begin{equation}
\mathbf{H}^{static} = \sum_{k=0}^{K} \Theta_k \Big(\alpha \mathbf{H} + (1-\alpha) \mathbf{A} \mathbf{H}\Big)
\end{equation}
where $\alpha \in (0,1)$ controls the residual contribution, $K$ is the propagation depth, and $\Theta_k$ are learnable parameters.

Dynamic Graph Structure (Wavelet Decomposition): To capture short-term or evolving dependencies, we perform multi-level wavelet decomposition (MWD) on the normalized input, i.e., $\mathbf{X}_{\text{norm}} = \sum_{l=1}^{L_d} \mathbf{X}^{(l)}, \; \mathbf{X}^{(l)} \in \mathbb{R}^{B \times L_l \times N}$, where $L_d$ denotes the decomposition level.

For each sub-sequence $\mathbf{X}^{(l)}$, a dynamic adjacency matrix is generated and graph convolution is applied:
\begin{equation}
\mathbf{A}^{dyn,(l)} = \text{Softmax}\Big(\tanh(W_1 \mathbf{X}^{(l)}) \cdot \tanh(W_2 \mathbf{X}^{(l)}) \Big)
\end{equation}
\begin{equation}
\mathbf{H}^{dyn,(l)} = \sum_{k=0}^{K} \Theta^{(l)}_k \Big(\alpha \mathbf{H}^{(l)} + (1-\alpha) \mathbf{A}^{dyn,(l)} \mathbf{H}^{(l)}\Big)
\end{equation}

The multi-scale dynamic graph feature set is denoted as:
\begin{equation}
\mathcal{H}^{dyn} = \{\mathbf{H}^{dyn,(1)}, \mathbf{H}^{dyn,(2)}, \dots, \mathbf{H}^{dyn,(L_d)}\}
\end{equation}

\subsection{Attention Gated Fusion}
\label{ssec:Attention}

To effectively integrate static and dynamic graph features, we employ an Attention Gated Fusion module. The input consists of the static graph representation $\mathbf{H}^{static} \in \mathbb{R}^{B \times D \times N}$ and the multi-scale dynamic graph set $\mathcal{H}^{dyn}$.  

All graph representations are stacked:
\begin{equation}
\mathcal{H} = [\mathbf{H}^{static}, \mathbf{H}^{dyn,(1)}, \dots, \mathbf{H}^{dyn,(L_d)}] \in \mathbb{R}^{B \times (L_d+1) \times D \times N}
\end{equation}

Each graph is pooled over nodes:
\begin{equation}
h_i = \text{AvgPool}_N(\mathcal{H}_i), \quad h_i \in \mathbb{R}^{B \times D}
\end{equation}

Attention weights are computed using a learnable query vector $q \in \mathbb{R}^D$:
\begin{equation}
\alpha_i = \frac{\exp(q^\top W_k h_i)}{\sum_j \exp(q^\top W_k h_j)}, \quad 
\mathbf{H}^{fusion} = \sum_{i=0}^{L_d} \alpha_i \mathcal{H}_i
\end{equation}

\subsection{Temporal Feature Learning}
\label{ssec:Temporal}
The fused representation $\mathbf{H}^{fusion} \in \mathbb{R}^{B \times D \times N}$ is processed by a multi-kernel dilated convolution Inception module to capture multi-scale temporal dependencies and includes a residual connection. The computation is formulated as:
\begin{equation}
\mathbf{H}_{concat} = \big[ \text{Conv1D}_{k,d}(\mathbf{H}^{fusion}) \big]_{k \in \{3,5\}, d \in \{1,2\}}
\end{equation}
\begin{equation}
\mathbf{H}_{out} = \text{Conv1D}_1(\mathbf{H}_{concat}) + \text{Conv1D}_1(\mathbf{H}^{fusion})
\end{equation}
\begin{equation}
\mathbf{H}^{temp} = \text{LayerNorm}(\mathbf{H}_{out})
\end{equation}

This module captures both short-term local patterns and long-term temporal dependencies, providing high-level temporal features for the subsequent prediction module output layer.

\begin{table*}[htbp]
  \centering
  \caption{Quantitative evaluation of time series forecasting models with $L = 96$ and prediction length $T \in \{96, 192, 336, 720\}$.}
  \resizebox{\textwidth}{!}{%
    \begin{tabular}{cc|cccccccccccccc}
    \toprule
    \multicolumn{2}{c}{Models} & \multicolumn{2}{c}{\textbf{Ours}} & \multicolumn{2}{c}{TimesNet} & \multicolumn{2}{c}{DLinear} & \multicolumn{2}{c}{MSGNet} & \multicolumn{2}{c}{MTGNN} & \multicolumn{2}{c}{Autoformer} & \multicolumn{2}{c}{PatchTST} \\
    \midrule
    \multicolumn{2}{c}{Metric} & MSE   & MAE   & MSE   & MAE   & MSE   & MAE   & MSE   & MAE   & MSE   & MAE   & MSE   & MAE   & MSE   & MAE \\
    \midrule
    \multicolumn{1}{c|}{\multirow{4}[2]{*}{\begin{sideways}Weather\end{sideways}}} & 96    & \underline{0.169} & \underline{0.215} & 0.172 & 0.220 & 0.196 & 0.255 & \textbf{0.163} & \textbf{0.212} & 0.171 & 0.231 & 0.266 & 0.336 & 0.184 & 0.227 \\
    \multicolumn{1}{c|}{} & 192   & \textbf{0.211} & \underline{0.256} & 0.219 & 0.261 & 0.237 & 0.296 & \underline{0.212} & \textbf{0.254} & 0.215 & 0.274 & 0.307 & 0.367 & 0.234 & 0.265 \\
    \multicolumn{1}{c|}{} & 336   & \textbf{0.269} & \textbf{0.297} & 0.280 & 0.306 & 0.283 & 0.335 & \underline{0.272} & \underline{0.299} & 0.266 & 0.313 & 0.359 & 0.395 & 0.284 & 0.301 \\
    \multicolumn{1}{c|}{} & 720   &\textbf{0.346} & \textbf{0.348} & 0.365 & 0.359 & \underline{0.345} & 0.381 & 0.350 & \textbf{0.348} & \underline{0.344} & 0.375 & 0.419 & 0.428 & 0.356 & \underline{0.349} \\
    \midrule
    \multicolumn{1}{c|}{\multirow{4}[2]{*}{\begin{sideways}ETT*\end{sideways}}} & 96    & \textbf{0.343} & \textbf{0.388} & \underline{0.384} & 0.402 & 0.386 & \underline{0.400} & 0.390 & 0.411 & 0.440 & 0.450 & 0.449 & 0.459 & 0.460 & 0.447 \\
    \multicolumn{1}{c|}{} & 192   & \textbf{0.361} & \textbf{0.407} & \underline{0.436} & \underline{0.429} & 0.437 & 0.432 & 0.442 & 0.442 & 0.449 & 0.433 & 0.500 & 0.482 & 0.512 & 0.477 \\
    \multicolumn{1}{c|}{} & 336   & \textbf{0.370} & \textbf{0.419} & \underline{0.438} & 0.469 & 0.481 & \underline{0.459} & 0.480 & \textbf{0.468} & 0.598 & 0.534 & 0.521 & 0.496 & 0.546 & 0.496 \\
    \multicolumn{1}{c|}{} & 720   & \textbf{0.454} & \textbf{0.486} & 0.521 & 0.500 & 0.519 & 0.516 & \underline{0.494} & \underline{0.488} & 0.685 & 0.620 & 0.514 & 0.512 & 0.544 & 0.517 \\
    \midrule
    \multicolumn{1}{c|}{\multirow{4}[2]{*}{\begin{sideways}Electricity\end{sideways}}} & 96    & 0.179 & \textbf{0.268} & \textbf{0.162} & \underline{0.272} & 0.197 & 0.282 & \underline{0.165} & 0.274 & 0.211 & 0.305 & 0.201 & 0.317 & 0.190 & 0.296 \\
    \multicolumn{1}{c|}{} & 192   & \textbf{0.181} & \textbf{0.270} & \underline{0.184} & \underline{0.289} & 0.196 & 0.285 & \underline{0.184} & 0.292 & 0.225 & 0.319 & 0.222 & 0.334 & 0.199 & 0.304 \\
    \multicolumn{1}{c|}{} & 336   & \underline{0.198} & \textbf{0.293} & \underline{0.198} & \underline{0.300} & 0.209 & 0.301 & \textbf{0.195} & 0.302 & 0.247 & 0.340 & 0.231 & 0.338 & 0.217 & 0.319 \\
    \multicolumn{1}{c|}{} & 720   & 0.248 & \underline{0.325} & \textbf{0.220} & \textbf{0.320} & 0.245 & 0.333 & \underline{0.231} & 0.332 & 0.287 & 0.373 & 0.254 & 0.361 & 0.258 & 0.352 \\
    \midrule
    \multicolumn{1}{c|}{\multirow{4}[2]{*}{\begin{sideways}Exchange\end{sideways}}} & 96    & \textbf{0.087} & \underline{0.207} & 0.107 & 0.234 & \underline{0.088} & 0.218 & 0.102 & 0.230 & 0.267 & 0.378 & 0.197 & 0.323 & \underline{0.088} & \textbf{0.205} \\
    \multicolumn{1}{c|}{} & 192   & \underline{0.181} & \underline{0.303} & 0.226 & 0.344 & \textbf{0.176} & 0.315 & 0.195 & 0.317 & 0.590 & 0.578 & 0.300 & 0.369 & \textbf{0.176} & \textbf{0.299} \\
    \multicolumn{1}{c|}{} & 336   & 0.342 & \underline{0.425} & 0.367 & 0.448 & \underline{0.313} & \textbf{0.427} & \underline{0.359} & 0.436 & 0.939 & 0.749 & 0.509 & 0.524 & \textbf{0.301} & \textbf{0.397} \\
    \multicolumn{1}{c|}{} & 720   & \underline{0.869} & \underline{0.705} & 0.964 & 0.746 & \textbf{0.839} & \textbf{0.695} & 0.940 & 0.738 & 1.107 & 0.834 & 1.447 & 0.941 & 0.901 & 0.714 \\
    \midrule
    \multicolumn{2}{c|}{$1^{st}$ Count} & \multicolumn{2}{c|}{\textbf{18}} & \multicolumn{2}{c|}{3} & \multicolumn{2}{c|}{4} & \multicolumn{2}{c|}{\underline{6}} & \multicolumn{2}{c|}{0} & \multicolumn{2}{c|}{0} & \multicolumn{2}{c}{5} \\
    \bottomrule
    \end{tabular}%
  }
  \label{tab:addlabel}%
\end{table*}%

\section{EXPERIMENTS}
\label{sec:EXPERIMENTS}
\subsection{Experimental Setup}
\label{ssec:subhead}

\noindent\textbf{Datasets.} To evaluate the effectiveness of our proposed method, we conducted experiments on seven benchmark time series datasets: ETT (h1, h2, m1, m2), Exchange-Rate, Electricity and Weather. For each dataset, we split the data into training, validation, and test sets with a 7:2:1 ratio.

\noindent\textbf{Baselines.} We evaluate our method against seven SOTA models, including GNN-based MSGNet \cite{Cai2024} and MTGNN \cite{wu2020connecting}, CNN-based TimesNet \cite{wu2022timesnet}, Transformer-based PatchTST \cite{nie2022time} and Autoformer \cite{wu2021autoformer}, and the MLP-based DLinear \cite{Zeng2023}.

\noindent\textbf{Implementation.} All experiments were implemented using PyTorch on a single NVIDIA GTX 4090 24GB GPU. The model was optimized with the Adam optimizer and $Batch=32$. Following standard time series forecasting protocols, the lookback length was set to $L = 96$, and $T \in \{96, 192, 336, 720\}$. Forecast performance was evaluated using Mean Squared Error (MSE) and Mean Absolute Error (MAE) ~\cite{Cai2024,  Zeng2023,  wu2021autoformer,wu2022timesnet}.

\subsection{Results and Analysis}
\label{ssec:subhead}

\begin{figure}[h]
\setlength{\abovecaptionskip}{2pt} 
\setlength{\belowcaptionskip}{2pt} 

\begin{minipage}[b]{0.49\linewidth}
  \centering
  \includegraphics[width=3.9cm]{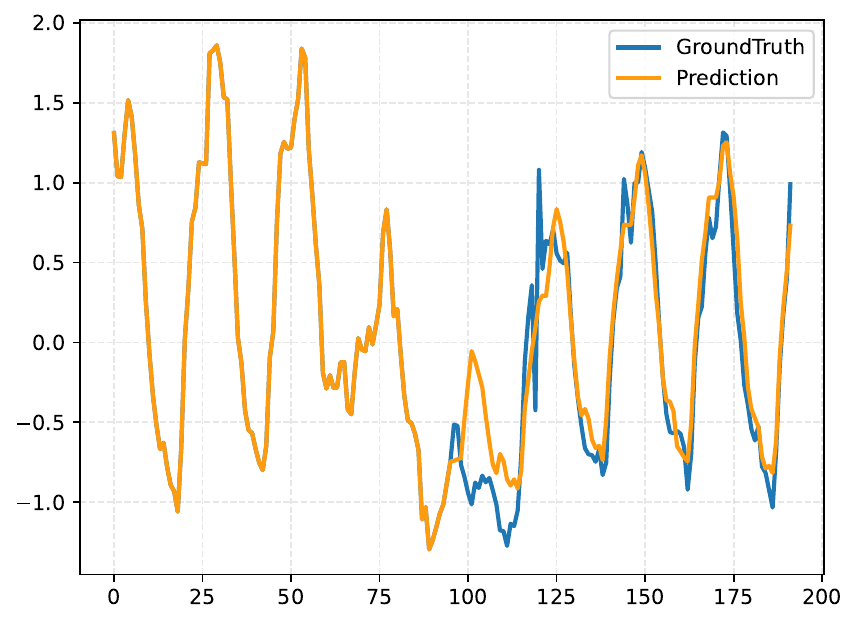}
  \centerline{(a) Electricity, output-96}
\end{minipage}
\hfill
\begin{minipage}[b]{0.49\linewidth}
  \centering
  \includegraphics[width=3.9cm]{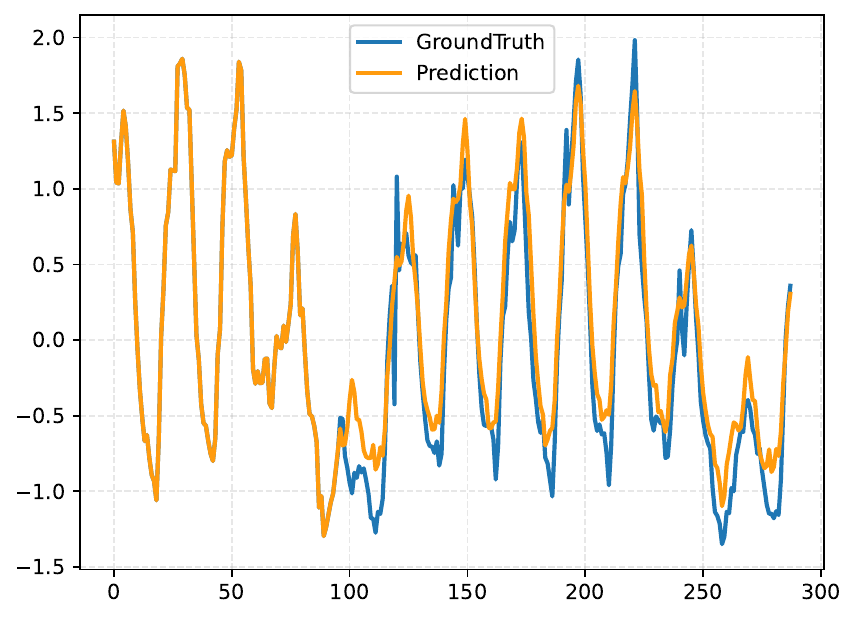}
  \centerline{(b) Electricity, output-192}
\end{minipage}

\begin{minipage}[b]{0.49\linewidth}
  \centering
  \includegraphics[width=3.9cm]{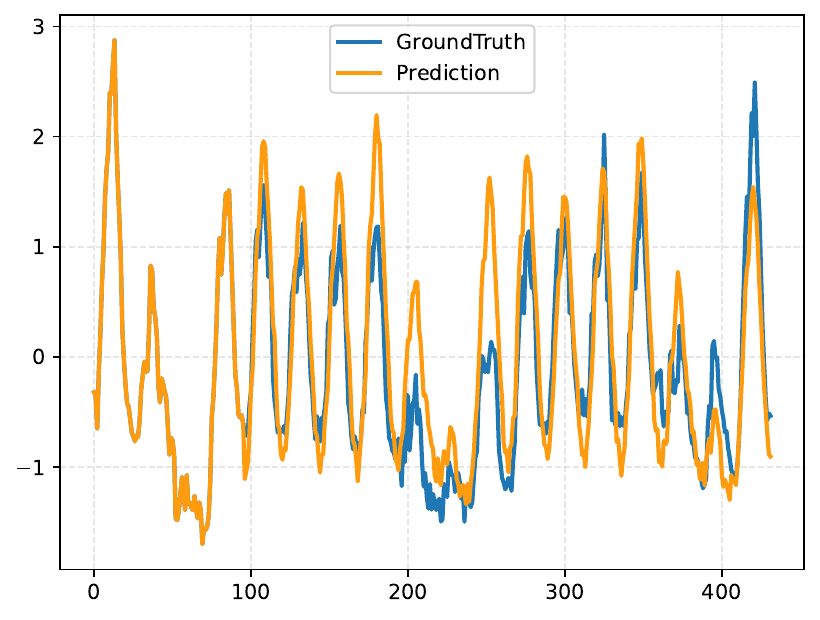}
  \centerline{(c) Electricity, output-336}
\end{minipage}
\hfill
\begin{minipage}[b]{0.49\linewidth}
  \centering
  \includegraphics[width=3.9cm]{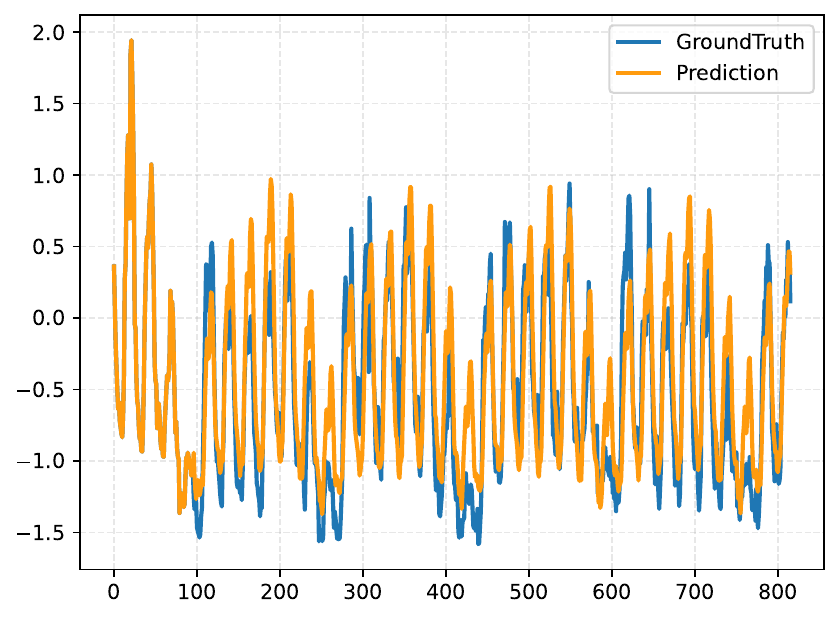}
  \centerline{(d) Electricity, output-720}
\end{minipage}

\caption{Visualization comparison of forecasting results.}
\label{fig:res}
\end{figure}

From the main results in Table~\ref{tab:addlabel}, our model achieves the best or second-best performance in most test tasks (best results in \textbf{bold}, second-best underlined), demonstrating its effectiveness and robustness. The advantage is particularly notable on the Weather and ETT* datasets, which exhibit pronounced multi-scale and periodic patterns. Specifically, on ETT*, our model attains the best MSE and MAE across all four prediction horizons. In comparison, GNN-based methods such as MTGNN and MSGNet perform worse, as they typically learn a single global static graph or fail to dynamically integrate graph structures with the inherent multi-scale characteristics of the signals, limiting their ability to capture evolving inter-variable dependencies. For visualization, partial prediction results are illustrated in Figure~\ref{fig:res}. Note that ETT* represents ETTh1, and other experimental results can be reproduced using the publicly available code.

\subsection{Ablation Study}
\label{ssec:subhead}

The ablation results on ETTh1, Exchange, and Weather (Table~\ref{tab:my_label}) demonstrate the importance of each component in SDGFNet. Removing graph structure learning (w/o-GSL) leads to a clear performance drop, highlighting the necessity of modeling both static and dynamic inter-series dependencies. Similarly, excluding the attention-gated fusion module (w/o-GF) degrades accuracy, verifying the effectiveness of integrating static and dynamic graphs. Moreover, removing temporal feature learning (w/o-TFL) also reduces performance, showing that multi-kernel dilated convolutions are essential for capturing cross-scale temporal dependencies. Overall, these results confirm that all three modules are indispensable to the superior performance of SDGFNet.

\begin{table}[htbp]
\centering
\caption{Ablation analysis of SDGFNet.}
\label{tab:my_label}
\resizebox{\columnwidth}{!}{
\begin{tabular}{ccccccc}
\hline
\textbf{Dataset} & \multicolumn{2}{c}{\textbf{ETTh1}} & \multicolumn{2}{c}{\textbf{Exchange}} & \multicolumn{2}{c}{\textbf{Weather}} \\ \hline
\textbf{Metric}  & \textbf{MSE}      & \textbf{MAE}      & \textbf{MSE}      & \textbf{MAE}      & \textbf{MSE}      & \textbf{MAE}      \\ \hline
SDGFNet  & \textbf{0.343} & \textbf{0.388}& \textbf{0.087} & \textbf{0.207}& \textbf{0.169} & \textbf{0.215}    \\
w/o-GSL     & 0.366      & 0.401    & 0.091         & 0.219        & 0.176       & 0.224    \\
w/o-GF      & 0.371      & 0.397     & 0.102       & 0.221     & 0.185             & 0.241     \\
w/o-TFL     & 0.368      & 0.409    & 0.099         & 0.215     & 0.181          & 0.237    \\
\hline
\end{tabular}}
\end{table}

\section{CONCLUSION}
\label{sec:foot}

In this paper, we introduced the SDGF Network, a novel dual-branch architecture for multivariate time series forecasting. By explicitly decoupling and fusing static global correlations with multi-scale dynamic dependencies learned via wavelet decomposition, our model captures a richer set of inter-series relationships. An attention mechanism intelligently integrates these complementary features. Extensive experiments on several benchmark datasets demonstrate the effectiveness and competitive performance of our proposed method.

\vfill\pagebreak

\bibliographystyle{IEEEbib}
\bibliography{strings,refs}

\end{document}